\documentclass{article}
\PassOptionsToPackage{numbers}{natbib}
\usepackage[final]{airov24} %
\usepackage{kpfonts}
\usepackage{graphicx} %
\usepackage{hyperref}
\usepackage{csquotes}
\usepackage{bm}
\usepackage{algpseudocode}
\usepackage{algorithm}
\usepackage{caption}
\usepackage{subcaption}
\title{Unsupervised Learning of Effective Actions in Robotics}
\author{
  Marko Zari\'{c}$^1$ \hspace{3pt} Jakob Hollenstein$^1$ \hspace{3pt} Justus Piater$^{1,2}$ \hspace{3pt} Erwan Renaudo$^1$\\
  $^1$Department of Computer Science, University of Innsbruck, Innsbruck, Austria \\
  $^2$Digital Science Center, University of Innsbruck, Innsbruck, Austria \\
  \texttt{firstname.lastname@uibk.ac.at} \\
}

\usepackage{color}

\newcommand{\actproto}{action prototypes}

\begin{document}

\maketitle

\begin{abstract} 
Learning actions that are relevant to decision-making and can be executed effectively is a key problem in autonomous robotics. Current state-of-the-art action representations in robotics lack proper effect-driven learning of the robot's actions. Although successful in solving manipulation tasks, deep learning methods also lack this ability, in addition to their high cost in terms of memory or training data. In this paper, we propose an unsupervised algorithm to discretize a continuous motion space and generate \enquote{action prototypes}, each producing different effects in the environment. After an exploration phase, the algorithm automatically builds a representation of the effects and groups motions into action prototypes, where motions more likely to produce an effect are represented more than those that lead to negligible changes. We evaluate our method on a simulated stair-climbing reinforcement learning task, and the preliminary results show that our effect driven discretization outperforms uniformly and randomly sampled discretizations in convergence speed and maximum reward.
\end{abstract}

\section{Introduction}

How to autonomously learn effective actions in robotics is a crucial question to enable robots to tackle diverse tasks in realistic environments with low supervision. The importance of action is highlighted in cognitive science, where recent developments propose to redefine cognition as \enquote{embodied action} \cite{EngelMKK2013}. On the other hand, implicit models of action are recurring in autonomous robotics, necessary for various systems in the robot: symbolic actions represented as operators allow long-term action planning \citep[chap 14]{SicilianoK2016}, continuous actions are encoded as motion primitives or policies in order to be executed in the environment \citep[chap 15]{SicilianoK2016}. Advanced perception capabilities like affordance detection process sensor data in regards to the actions the robot is able to perform in the environment \citep{SahinCDUU2007, ZechHRRUP2017}.

However, in robotics, this diversity of action models at several levels of abstraction leads to partial and heterogeneous representations, a problem that is identified by \cite{ZechRHZP2019}. They propose a tentative definition of action in robotics in order to formalize the design and learning of action representation. Supported by results from other fields of cognitive science, they particularly emphasize the key importance of the concept of \emph{effect} produced by motor behaviors, in order to define an action. They also examine how researchers approach the question of action representations in robotics and highlight several open challenges to be addressed. In this work, we focus on addressing one of them: \enquote{Intensifying effect-centricity and effect grounding} (in action representation). More specifically, we are interested in providing discrete actions often referred to as \emph{action symbols} at the decision level that are \emph{grounded in the actual physical effects produced by the interactions of the robot with its environment}.

For that, we design an algorithm to generate what we call \enquote{\actproto{}s}: each prototype is a set of motion parameters that let the robot produce a specific type of effect. Generating these prototypes involves a) finding out which types of effects are possible in a given environment through exploration, b) building the class of equivalent effects, and c) finding a set of representative motion parameter sets that allows achieving such effects reliably. Our work contributes by formalizing and implementing an effect-centric action space discretization algorithm\footnote{https://github.com/marko-zaric/action-prototype-gen.git}. Additionally, we create a toy environment in Gazebo with a Gym-Wrapper called \enquote{Up The Stairs}\footnote{https://github.com/marko-zaric/up-the-stairs.git} with a continuous motion and observation space well suited for action space discretization evaluations. Finally, we perform a comparative evaluation of the proposed effect-centric discretization algorithm and trivial discretization approaches on the \enquote{Up The Stairs} environment.

The rest of this paper is organized as follows:  Section \ref{sec:rel_work} reviews existing approaches that learn \actproto{}, Section \ref{sec:methods} describes our multi-step approach relying on unsupervised methods to build effect and action representations. In Section \ref{sec:results}, we describe the environment in which we evaluate this method as well as its relevance for decision-making in robotics in a reinforcement learning (RL) problem. Finally, we examine the implications and potential limitations of the method in Section  \ref{sec:discussion}.

\section{\label{sec:rel_work}Related Work}

In robot action learning, a large part of the research effort is dedicated to learning (encoding of) trajectories useful to solve a task, as shown by the several surveys on robot manipulation \citep{GamsPNU2022, KroemerNK2021}. Standard representations include motion primitives, as often done in learning by demonstration, and policies learned with reinforcement learning. To the best of our knowledge, the explicit learning of actions based on the effect they produce is not often explicitly addressed \citep{ZechRHZP2019}. The current state-of-the-art approach to learning motions in robotics is to train an end-to-end deep neural network through reinforcement learning \citep{SaeedNRA2021,WangPLWJ2022,HanMSC2023}. However, limitations include the number of parameters, training data, and time required for the learning process to converge \citep{LevineFDA2016}.  

A promising approach proposed by \cite{ZademMN2023} uses a hierarchical reinforcement learning approach to learn both a goal abstraction and an agent's policy for a given task. They show that the goal representation (i.e., the policy effects) learned by their algorithm is necessary to solve tasks in complex continuous environments with sparse rewards. However, they learn to solve a task in a RL setting, which means their representations are tied to a task and biased by the reward function.

The dependency on a reward function is lifted when explicitly considering the effects: one can find effect-based action learning in the affordance learning literature in robotics. The focus is on learning the relationship between environmental features, robot behavior, and their corresponding effects \citep{SahinCDUU2007}. \cite{ChavezLC2016} use a Bayesian representation of the causal relation between the environment, the robot actions, and the effects. They start with pre-coded elementary actions and handcrafted effect detectors and can find affordances based on collected data but do not let the robot discover effects. \cite{AndriesCCGG2018} propose an affordance equivalence operator based on the produced effect. Using a Bayesian Network, they model relations between object, action, and additionally, the actor from experimental data. These models allow to find equivalent elements, especially equivalent action to obtain the same effect, but do not modify the set of available actions. 

A step towards learning a relevant effect representation is made by \citep{UyanikCBYKS2013} in the context of social affordances. Despite using pre-coded actions, they compute effect codes based on the continuous variations of the features in effect space. This representation produces effect equivalence classes based on the mean and standard deviation of effects produced by one action. No supervision is required, and effect classes are thus autonomously discovered. The limitation here is that actions being hard-coded, some have multiple outcomes, contrary to the idea that the effect defines the action.

\section{\label{sec:methods}Methods}
Solving the underlying problem for continuous systems is more challenging due to the infinite number of possible actions requiring parametric functions to describe action distributions. The primary aim of action space discretization is to maintain the simplicity of discrete actions in continuous control problems \cite{tang2020discretizing}. To tackle these tasks in a sample efficient manner, we consider a developmental scenario where the robot can explore interacting with the environment. The agent can issue commands in continuous motion space, and the goal is to find discrete action prototypes in a sample-efficient way where each prototype performs a reliable effect in the environment. The exploration phase is divided into three main stages: motion sampling, effect region clustering, and action prototype generation. All three stages operate unsupervised, resulting in ready-to-use action prototypes for a downstream discrete RL algorithm.

\subsection{Motion sampling}
The initial setting for this method is a robot environment with a continuous state space $\mathcal{O}$ defined as $\mathcal{O} = \varprod_{i=1}^n O_i$ where each $O_i$ is a bounded interval for the $i$th feature value ($\varprod$ denotes the Cartesian Product). We consider a continuous motion space $\mathcal{M} = \varprod_{i=1}^n M_i$ where $M_i$ takes values in a bounded interval. These are the motor controls in robot actuators with their respective minimal and maximal value. We define an effect $e_t$ at time step $t$ as follows 
\begin{equation}
    e_t = s_t - s_{t-1} \quad  \forall t \, : \, s_t \in \mathcal{O}
\end{equation}
where $\mathcal{O}$ is an observation space and $s_t$ and $s_{t-1}$ are two consecutive states linked by motion $m_{t-1}$. At this stage, the robot samples a random motion $m_t$ uniformly from motion space $\mathcal{M}$ at each step and performs it always starting from the environment's initial position. When performing a motion, our method focuses on the effect it produces in the environment and stores the resulting $(m_{t-1},\, e_t)$ tuples in $\Omega$. Pseudocode for this stage is given in Algorithm \ref{alg:motion-sampling}. Since the environment is reset after every collected sample, we drop the time index $t$. 

\begin{algorithm}[t]
\caption{Motion sampling}\label{alg:motion-sampling}
\begin{algorithmic}
\Require $\mathcal{O} = \varprod_{j=1}^n O_j$ and $\mathcal{M} = \varprod_{j=1}^n M_j$ \Comment{Motion space $\mathcal{M}$ and Observation space $\mathcal{O}$}
\State \hspace{-12pt} \textbf{Return:} $\Omega$ \Comment{Collection of (motion, effect) tuples}
\State $i \gets 0$
\While{$N \geq i$}
\State Initialize $s$ \Comment{Reset environment at each sample}
\State Sample $m_i \sim \mathrm{U}(\mathcal{M})$
\State Perform motion $m_i$ and observe $s'$
\State $e_i = s' - s$
\State Store $(m_i, e_i)$ to $\Omega$
\State $i \gets i + 1$
\EndWhile
\end{algorithmic}
\end{algorithm}

\subsection{Effect region clustering}
Effect region clustering is achieved by grouping the effects resulting from the sampled motions to generate effect classes $\mathcal{C}_k$, as described by Algorithm \ref{alg:effect-clust}. 
\begin{algorithm}[t]
\caption{Effect region clustering}\label{alg:effect-clust}
\begin{algorithmic}
\Require $N$ motion-effect tuples $\Omega$, $\psi$ max clusters
\State \hspace{-12pt} \textbf{Return:} $\mathcal{C}_1, \ldots , \mathcal{C}_N$ \Comment{$N$ distinct Effect Regions}
\If{dim$(e) > 1$}
\State $\theta \gets 0$ \Comment{maximal silhouette score}
\State $\phi \gets 2$ \Comment{best number of clusters}
\For{\textbf{each} $i \in [2, \ldots, \psi]$}
\State $s \gets $ silhouette score of $\mathrm{KMEANS}(\{e \in \Omega\},\mathrm{clusters}=i)$
\If{$\theta < s$}
\State $\theta \gets s$ 
\State $\phi \gets i$ 
\EndIf
\EndFor
\State $\{\mathcal{C}_k : k \in [1, \ldots \phi]\} \gets \mathrm{KMEANS}(\{e \in \Omega\},\mathrm{clusters}=\phi)$
\Else 
\State $\mathrm{borders}, \mathrm{bins} \gets \mathrm{Histogram}(\{e \in \Omega\})$
\State Create $\mathcal{C}_k$ for non-empty bins  
\For{\textbf{each} $(e,m) \in \Omega$}
\State Append $(e,m)$ to $\mathcal{C}_k$ where $e \in \mathrm{borders}(\mathcal{C}_k)$
\EndFor
\EndIf
\State 
\end{algorithmic}
\end{algorithm}
A different method for grouping the samples into effect categories is selected depending on the specified number of relevant effect dimensions. If only one effect dimension is interesting for the task, simple histogram binning groups the samples into the respective categories. Otherwise, we opt for the K-Means algorithm (\cite{kmeans}) to cluster effects in multidimensional space. In order to find the best number of classes to represent the data, we perform multiple iterations of K-Means while increasing the number of generated centroids. The iteration that produces clusters with the highest silhouette score are considered to be the current best representation of possible effects and used as $\mathcal{C}_k$. In each version, labels are amended for each sample generated at the previous stage, resulting in $(m_{t-1},\, e_t,\, k)$ tuples.

\subsection{Action prototype generation}

Each class $\mathcal{C}_k$ constitutes a collection of similar effects, each with a respective motion that caused the effect. This one-to-one relationship allows us to group the sampled motions by their associated effect label $e_k$. We call this collection of motions \enquote{action} $\mathcal{A}_k$: all the motions that achieve the same class of effect.

In order to achieve $e_k$, any motion from action $\mathcal{A}_k$ can be performed. However, instead of maintaining all these individual motions, we want to find a small number of $p_k \in \mathcal{M}$ for each effect class $\mathcal{C}_k$. All $p_k$ are action prototypes available at the discrete decision-making level of the robot.

We selected the RGNG (Robust growing neural gas) algorithm by \cite{rgng} to generate the action prototypes. This algorithm utilizes a combination of topological learning and outlier resilience, which stays true to the underlying sample topology of the input data. 

RGNG needs a maximally allowed number of nodes on initialization. We consider an effect-driven approach to calculate this number based on the underlying effect samples of class $\mathcal{C}_k$. For each class $k$, we determine the mean $\mu_k^e$ and the standard deviation $\sigma_k^e$ in effect space. After normalizing the mean and standard deviation across all classes, we calculate the number of prototypes $\xi_k$ for each class according to Equation (\ref{eq:metric}).
\begin{equation}
    \xi_k = \Bigl\lfloor \frac{(1 - \textrm{cv}_k)\cdot \textrm{std}_k}{\min_k((1 - \textrm{cv}_k)\cdot \textrm{std}_k)} \Bigl\rfloor
\label{eq:metric}
\end{equation} 
\begin{equation}
    \textrm{cv}_k = \frac{\sigma^e_k}{\mu^e_k}
\label{eq:cv}
\end{equation} 
where $\textrm{cv}_k$ is the coefficient of variation (defined in Equation (\ref{eq:cv})). This is a statistical measure of dispersion relative to the mean. A high coefficient of variation means the action is less robust; therefore, the first term reduces the number of prototypes for unreliable actions. The second term reduces the number of prototypes for potentially reliable actions with little variability to avoid motion overlap. The division by the minimal value ensures that each encountered class gets at least one prototype. Finally, we run either RGNG or calculate the motion space mean $\mu_k^m$ to generate the action prototypes $p_k$ for each class $\mathcal{C}_k$

\begin{equation}
    p_k =  \begin{cases}
        \mu^m_k & \text{if } \; \xi_k == 1\\
        \textrm{RGNG}(\xi_k) & \text{if } \; \xi_k > 1.
    \end{cases}
\end{equation}

\section{\label{sec:results}Results}
\begin{figure}
    \centering
    \begin{subfigure}[b]{.49\textwidth}
         \centering
         \includegraphics[width=.8\textwidth]{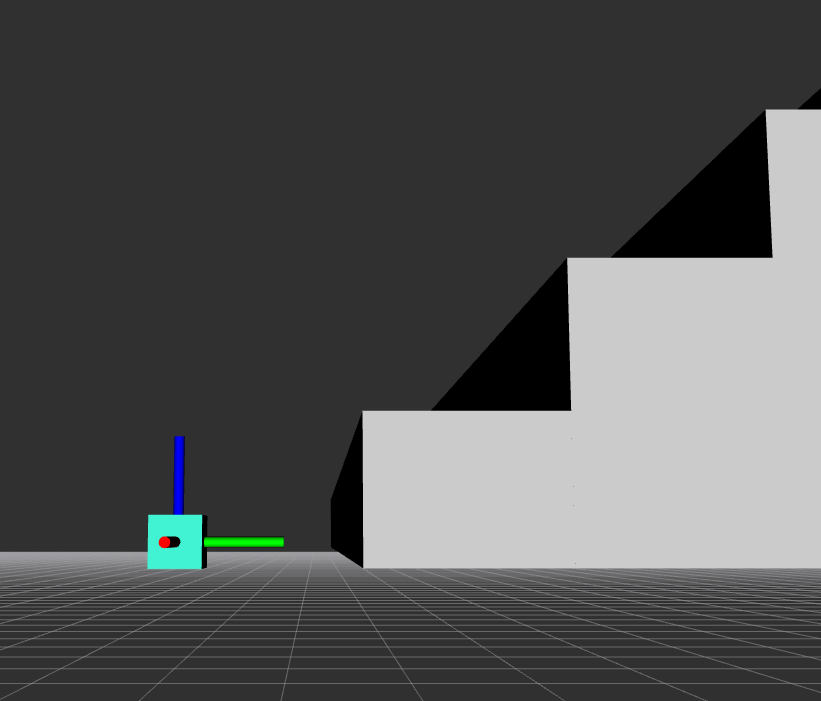}
         \caption{Initial position}
         \label{fig:start}
     \end{subfigure}
     \hfill
     \begin{subfigure}[b]{0.49\textwidth}
         \centering
         \includegraphics[width=.8\textwidth]{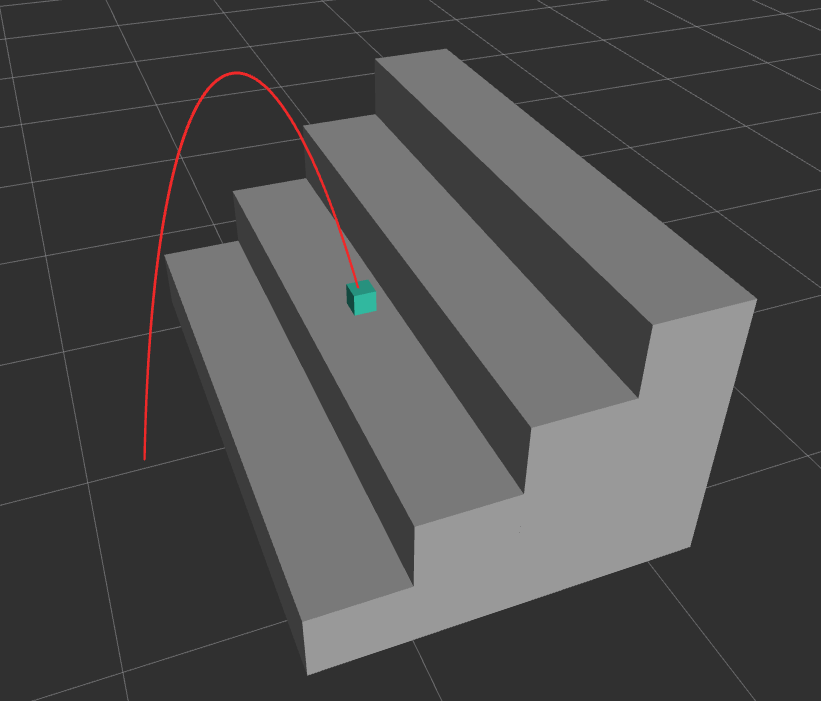}
         \caption{Performed jump}
         \label{fig:jump}
     \end{subfigure}
    \caption{The simulated environment: (a) the robot (the turquoise cube) starts facing a stairway consisting of 4 steps in its initial position. It can apply a force in the y-z plane to its center of mass (red: x, green: y, blue: z) (b) the robot after performing a motion, in its final position. The red line is the performed trajectory.}
    \label{fig:sim_env}
\end{figure}

We evaluate the algorithm in a simulated environment where a simple robot can move by \enquote{jumping}, i.e., applying a force to its center of mass. Figure \ref{fig:sim_env} illustrates the setup. The \enquote{jumping} command is
\begin{equation*}
m = \{ \bm{\alpha}, \mu \}
\end{equation*}
where $\bm{\alpha} = (0, \alpha, 0)$ is the direction vector of the force applied at the robot center of mass, and $\mu$ is its amplitude. The robot receives
\begin{equation*}
s = \{x, y, z, qx, qy, qz, qw, d_{obs}\}
\end{equation*}
as features, where x, y, z encodes its position in the world frame,  qx, qy, qz, qw is the quaternion representing its orientation and $d_{obs}$ the distance to the next obstacle on the x-axis.

\begin{figure}
    \centering
    \includegraphics[width=.5\textwidth]{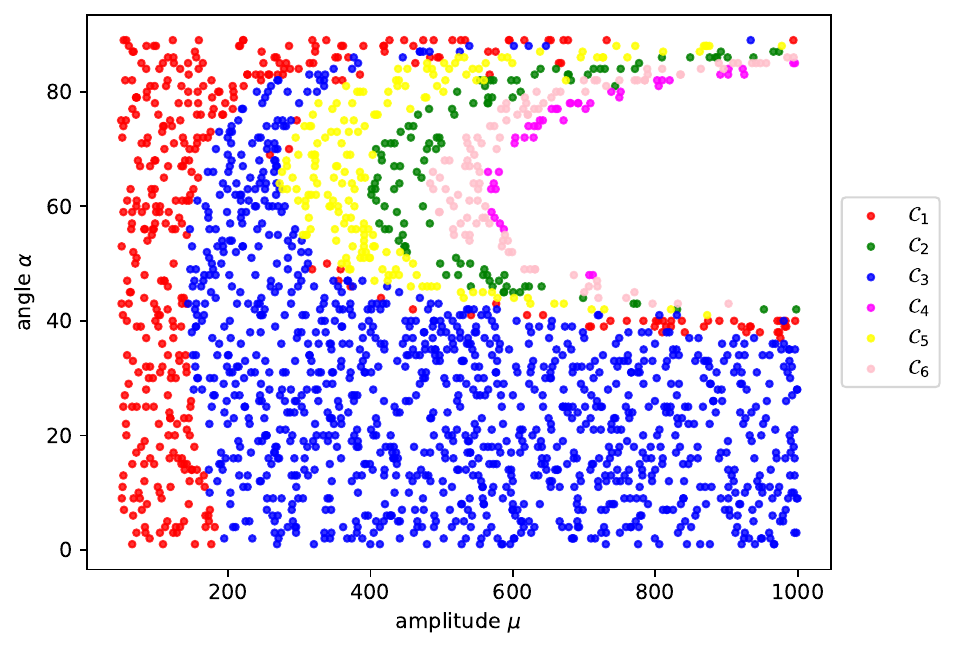}
    \caption{The visualization of the k-Means effect region clustering in y and z space in action space shows coherent classes even though the clustering was performed in effect space, which reaffirms these two spaces' correlation. Classes $\mathcal{C}_1$ and $\mathcal{C}_3$ correspond to no strict change in the environment where the robot does not reach the first step. The other regions in the top half represent a one-, two-, three-, or four-step height gain (from left to right).}
    \label{fig:motion_exploration}
\end{figure}

Figure \ref{fig:motion_exploration} shows the k-means clustering result on the collected data in effect space visualized in motion space. The method is able to find meaningful classes on the y,z space. Other dimensions of the effect space are random in this setting since the control action is applied in the y-z plane. Classes $\mathcal{C}_1$ and $\mathcal{C}_3$ correspond to no strict change in the environment where the robot does not reach the first step. In classes $\mathcal{C}_2$, $\mathcal{C}_4$, $\mathcal{C}_5$, $\mathcal{C}_6$, significant changes in the position of the robot are produced. Each subsequent class (in order $\mathcal{C}_5$, $\mathcal{C}_2$, $\mathcal{C}_6$, $\mathcal{C}_4$) represents the robot making it one step higher. The empty area in the top left of Figure \ref{fig:motion_exploration} corresponds to overshooting off the stairs, which we excluded in the final clustering.

\begin{figure}
    \centering
    \begin{subfigure}[b]{.49\textwidth}
         \centering
         \includegraphics[width=\textwidth]{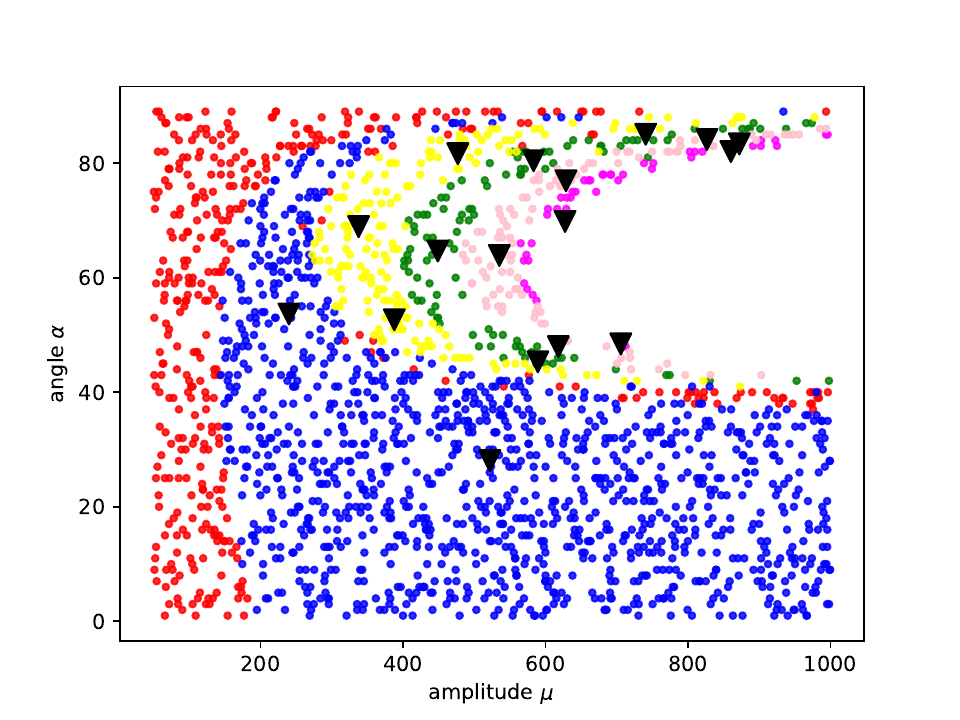}
         \caption{Effect Region Clustering + RGNG with variation dependent prototype quantity (Ours)}
         \label{fig:ourprot}
     \end{subfigure}
     \hfill
     \begin{subfigure}[b]{0.49\textwidth}
         \centering
         \includegraphics[width=\textwidth]{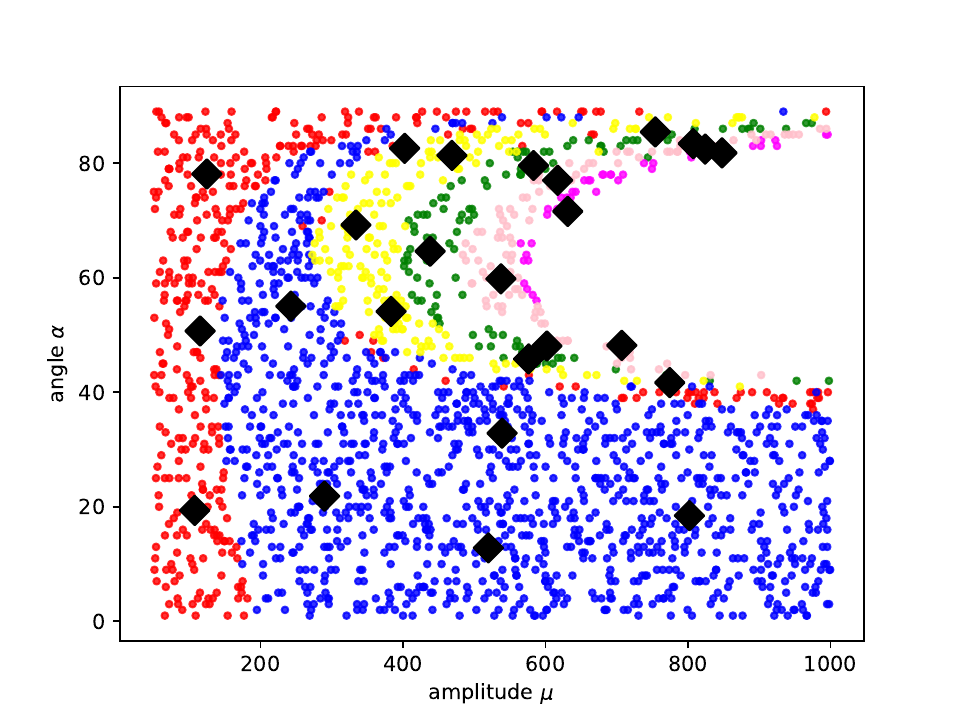}
         \caption{Effect Region Clustering + RGNG and fixed number of prototypes for all classes}
         \label{fig:rgngprot}
     \end{subfigure}
     \vfill
     \begin{subfigure}[b]{0.49\textwidth}
         \centering
         \includegraphics[width=\textwidth]{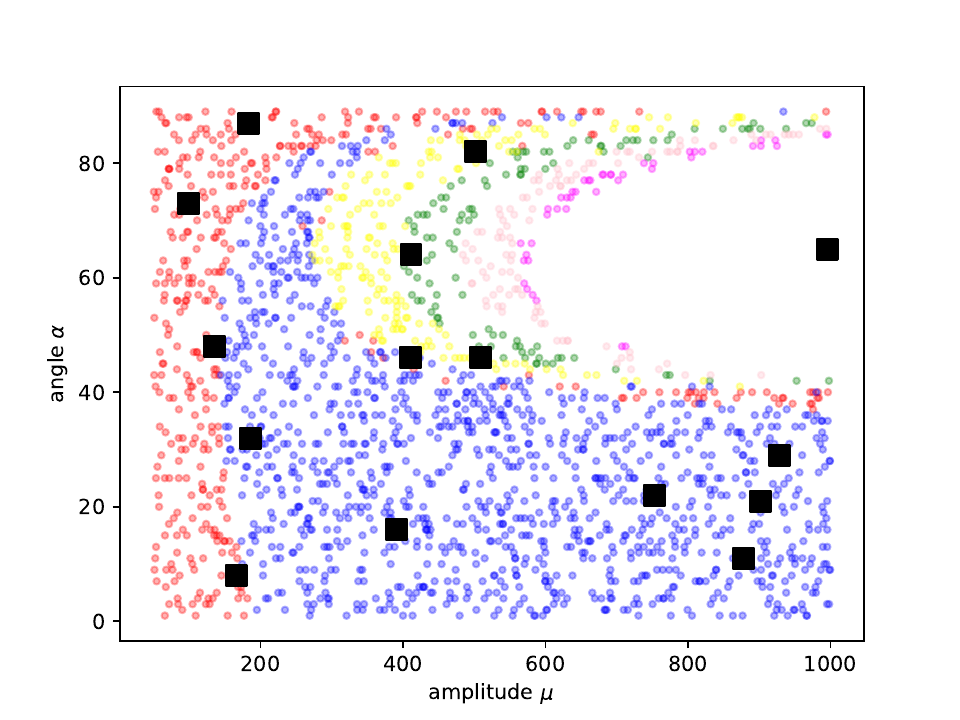}
         \caption{Random prototypes}
         \label{fig:randomprot}
     \end{subfigure}
     \hfill
     \begin{subfigure}[b]{0.49\textwidth}
         \centering
         \includegraphics[width=\textwidth]{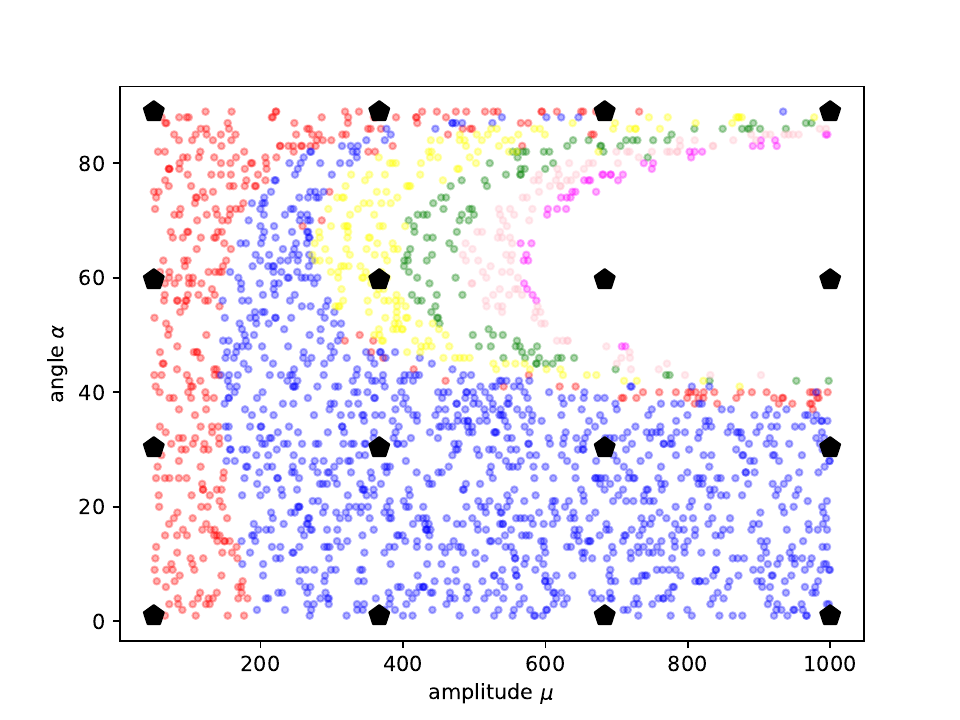}
         \caption{Uniform grid prototypes}
         \label{fig:uniformrandomprot}
     \end{subfigure}
    \caption{Action prototypes (visualized as regular polygons) found by (a) Effect Region Clustering with RGNG (Robust Growing Neural Gas Algorithm) and variation metric (Equation \ref{eq:metric}) for selecting the prototype quantity per class, (b) Effect Region Clustering with five prototypes per cluster RGNG, (c) random prototypes and (d) uniform grid prototypes. Methods (a) and (b) identify key change areas using Effect Region Clustering to generate prototypes across various effect motions. Method (b) generates excessive prototypes in stable areas, undermining the advantage of effect classes. Random and uniform methods yield many irrelevant prototypes. The colors of effect classes serve only as visual cues and do not affect methods (c) and (d).}
    \label{fig:motion_prototypes}
\end{figure}

Figure \ref{fig:motion_prototypes} shows the action prototype found by our method/other methods. Our method manages to find distinct prototypes in significant effect areas. RGNG without the metric in Equation (\ref{eq:metric}) finds distinct prototypes in significant effect areas but creates too many prototypes in "low effect" and "low variation in effect" areas because all areas are deemed as equally important. The uniformly random prototype selection fails to capture a large portion of the high-effect areas and cannot follow the underlying topology in the effect space.

To measure the quality of the generated prototypes, we constructed a reinforcement learning problem in the aforementioned environment with 15 steps instead of four. We created a custom Gymnasium \citep{towers_gymnasium_2023} environment wrapper for our simulation with reward function $r_t$ (from now on called "Up the stairs" environment). This reward function rewards jumping one step at a time and punishes falling proportional to the number of steps fallen. 

\begin{equation}
    r_t =  \begin{cases}
        1 & \text{if } \; (s_{t + 1}^z - s_{t}^z) >  0 \\
        -(s_{t + 1}^z - s_{t}^z)/0.3 & \text{if } \; (s_{t + 1}^z - s_{t}^z) <=  0.
    \end{cases}
\label{eq:reward}
\end{equation}

\begin{figure}[ht]
  \centering
  
  \begin{subfigure}[b]{0.8\textwidth} %
    \includegraphics[width=\textwidth]{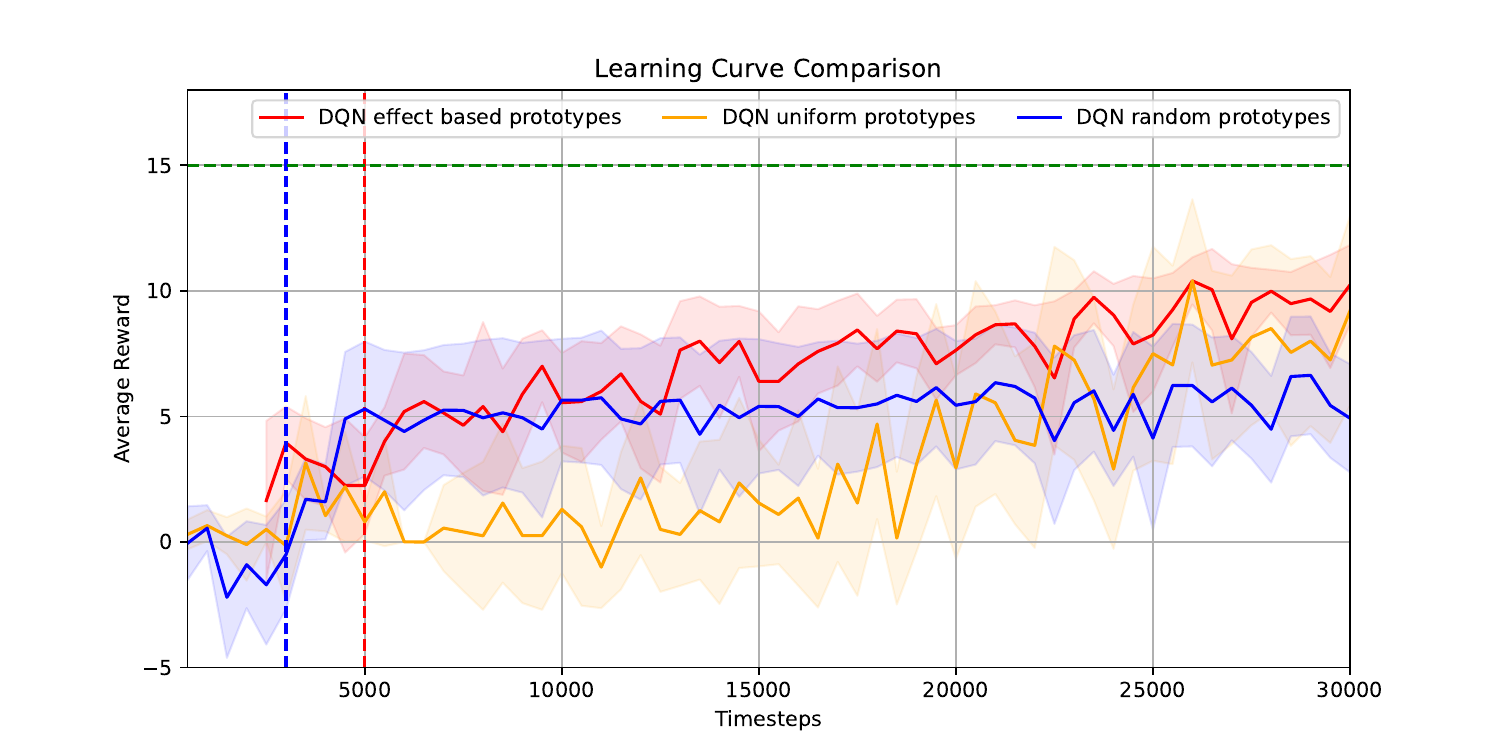} %
    \label{fig:sub1}
    \vspace{-10pt}
    \caption{}
  \end{subfigure}
  \vfill %
  \begin{subfigure}[b]{0.8\textwidth} %
    \includegraphics[width=\textwidth]{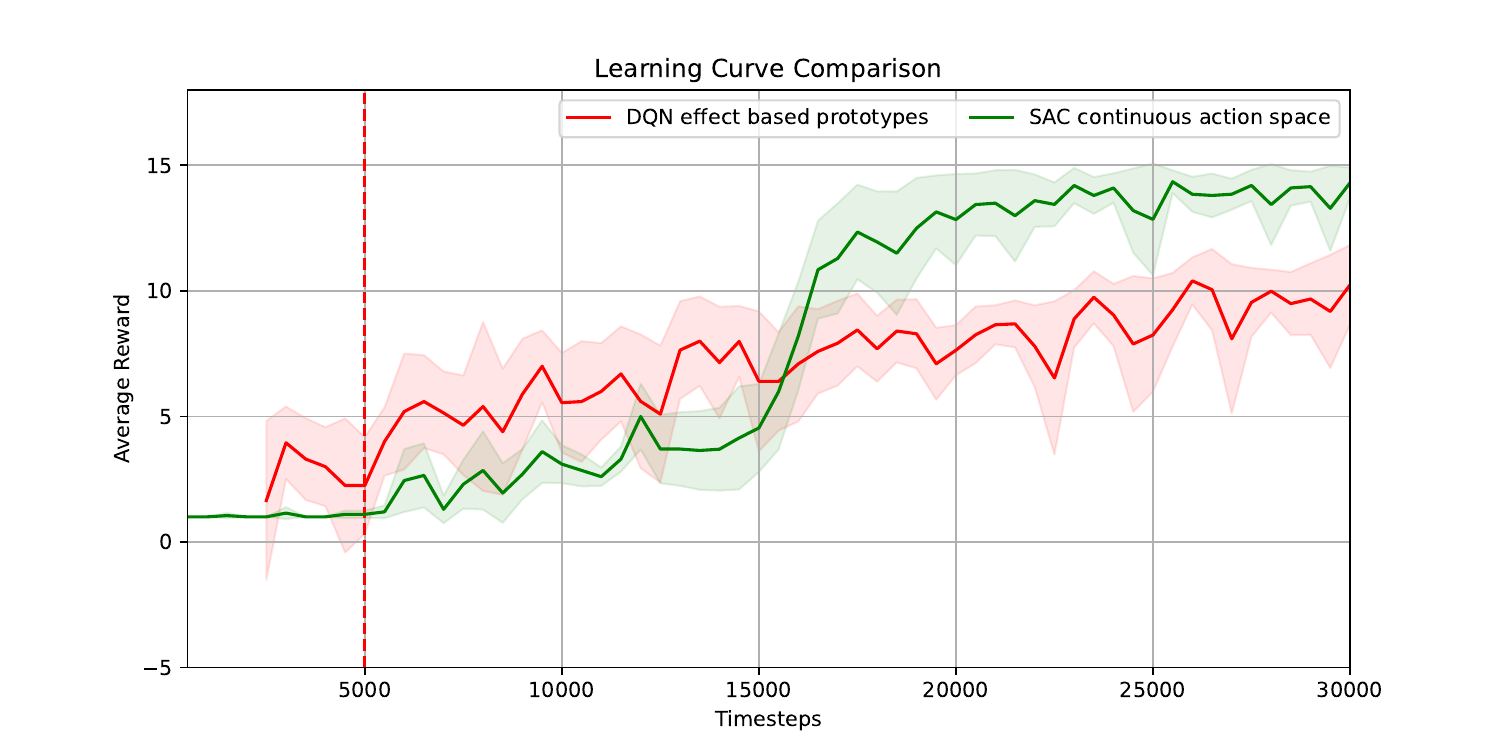} %
    \label{fig:aaaa}
    \vspace{-10pt}
    \caption{}
  \end{subfigure}
  \caption{During the "Up The Stairs" task learning, our effect-centric method, adjusted for 2000 samples for prototype generation, achieved faster convergence and higher maximal reward than random or uniform prototype generation methods. The SAC baseline, representing solvability in a continuous action space, has its maximal reward marked by a green dashed line at 15 in the graph (a). Graph (b) compares the learning curves of the SAC baseline, our effect-based discretization with DQN.}
  \label{fig:rl_performance}
\end{figure}

We used the SAC and the DQN implementation of Stable-Baselines3 \citep{stable-baselines3} as reinforcement learning agents in our evaluation. As a baseline for the achievable performance on this task, we trained the SAC agent \citep{sac} with 475146 parameters on the continuous motion space for 30000 steps. Additionally, we trained three Deep Q-learning agents with 5520 parameters each on our effect-based, uniform, and random prototypes. Figure \ref{fig:rl_performance} shows the mean reward with its respective standard deviation for each agent evaluated five times for each of the four seeds after every 500 timesteps. Each of the DQN agents collected data for 3000 steps into the replay buffer before starting to learn. The blue vertical dashed line in Figure \ref{fig:rl_performance} marks the point of learning start for the uniform and random prototypes. The red vertical dashed line is the sum of the 3000 DQN exploration steps and the 2000 motion samples collected before prototype generation.

Our method exhibits robustness similar to SAC regarding standard deviation, which the other methods fail to achieve. While all methods fall short of the performance achievable in the continuous setting with SAC, our method achieves the highest mean return out of the three discrete methods. The effect-based method holds advantages over both alternative discretization methods: Random prototypes, compared to effect-based prototypes, rise to their maximum performance similarly fast but fail to reach the same average reward. Uniform prototypes are close in average reward to our method but converge way slower. Finally, although the performance of our method could be better compared to SAC, it achieves it with an 85 times smaller number of parameters. Some runs of our method achieve the maximal performance, but more investigation is needed to understand the reason for the lower average. 

\section{\label{sec:discussion}Discussion}

Our custom "Up the stairs" environment shows promising first results in the search for automatic effect-based action discovery. A definite upside is that with a disproportionately smaller network size compared to the baseline SAC (factor 85), our method achieved competitive results with its preemptive effect-based prototype generation. One current limitation is the need for discrete effect spaces for clear effect class separation. Discrete effect spaces manifest through fixed boundary conditions in the environment (i.e., immovable stairs), which cause the effect class discovery to form clear boundaries as in Figure \ref{fig:motion_exploration}. There are no emergent effect category borders in purely continuous effect spaces (i.e., free space navigation). In future work, we want to investigate if effect-based discretization still yields benefits over the simpler uniform prototypes. Appendix B provides a visual aid for understanding the difference between discrete and continuous effect spaces.

When selecting the set of features that the algorithm should consider when evaluating the effects, there is a trade-off between performance and generalization. With each added available feature, the dimensionality of the clustering input space increases. If a lot of the added features are random, previously separate effect categories can merge into one. A visual representation of that can be found in Appendix C. If there is previous knowledge of insignificant effect features, removing them from consideration when clustering leads to a more robust effect class discovery. In future work, we want to investigate unsupervised options for measuring the randomness of each effect space feature so that the sweet spot of performance and generalization is discovered automatically.

The action prototype selection method is critical in enabling the robot's initial performance. Real-world settings and environments often do not exhibit evenly distributed high-effect regions. Our metric provides a reasonable estimate of how many prototypes each category should hold. After this initial decision, it is crucial to spread the action prototypes far from each other inside an effect category to create a wide range of different-looking motions in the robot's repertoire.  

\section{\label{sec:conclusion}Conclusion}

We presented a new effect-centric approach to learn a discrete action set that can be used by decision-making components of robots grounded in their actual effects in the environment. This approach only makes assumptions about the state dimensions where effects should be measured and does not rely on the design of a reward function. Our method is constructed task-agnostic and shows promising preliminary results: in our toy environment \enquote{Up The Stairs} effect-based action space discretization outperforms uniformly and randomly sampled action space discretization in convergence speed and maximum reward.  

\section*{Acknowledgements}
This research was funded in whole or in part by the Austrian Science Fund (FWF) [ELSA, DOI: 10.55776/I5755]. For open access purposes, the author has applied a CC BY public copyright license to any author accepted manuscript version arising from this submission.

\bibliographystyle{plainnat}
\bibliography{references}

\newpage

\appendix

\section*{APPENDIX}

\section{Action prototypes}
\begin{figure} [h]
    \centering
    \begin{subfigure}[b]{.49\textwidth}
         \centering
         \includegraphics[width=\textwidth]{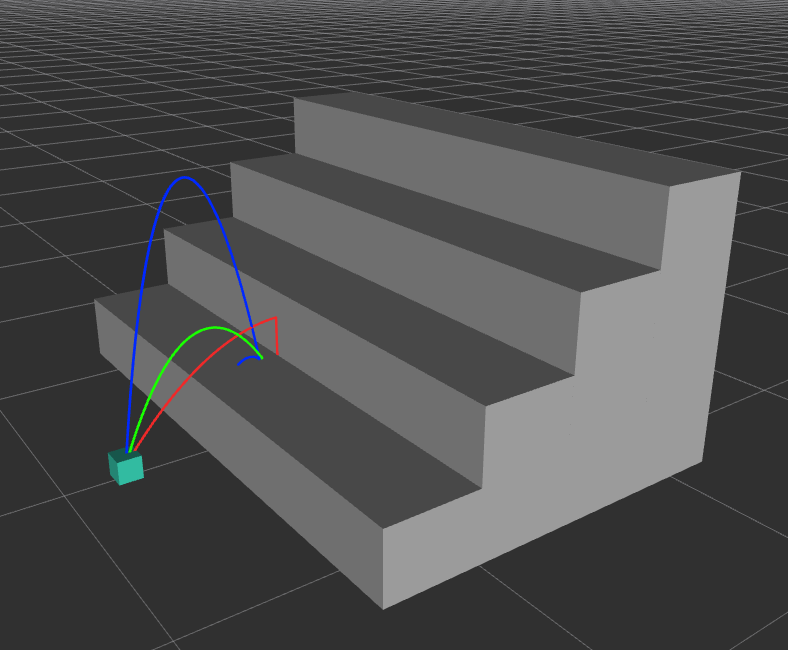}
         \caption{Action prototypes for action $\mathcal{C}_5$.}
     \end{subfigure}
     \hfill
     \begin{subfigure}[b]{0.49\textwidth}
         \centering
         \includegraphics[width=\textwidth]{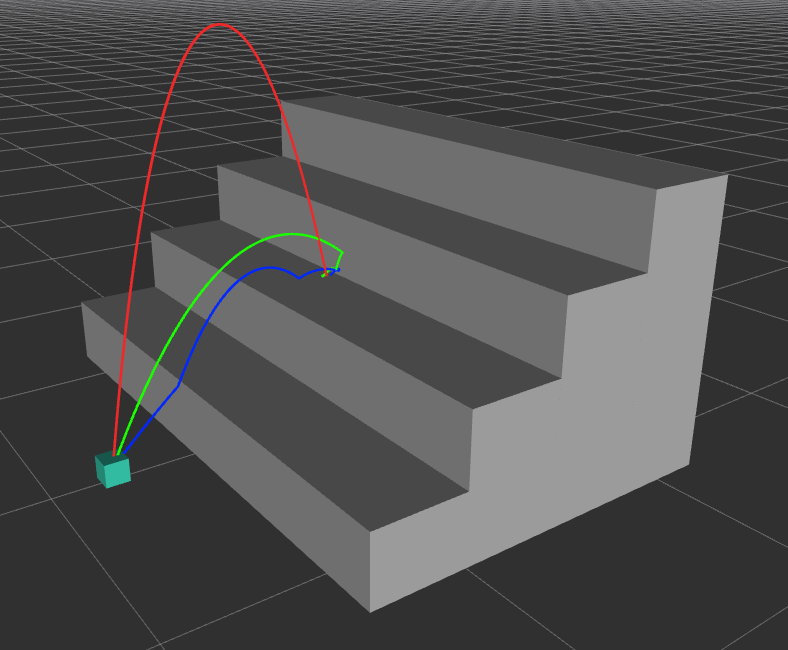}
         \caption{Action prototypes for action $\mathcal{C}_2$.}
     \end{subfigure}
    \caption{Figure (a) shows three action prototypes for the action "jump one step". Figure (b) shows three action prototypes for the action "jump two steps".}
    \label{fig:proto_example}
\end{figure}

This section shows three of the generated action prototypes using our effect based method for effect categories $\mathcal{C}_5$ and $\mathcal{C}_2$. In figure \ref{fig:proto_example} image (a) displays the trajectories of three action prototypes for the action a human would label "jump one step" and image (b) for the action "jump two steps". Here the strengths of the topological RGNG algorithm are clearly visible as each action prototype has a significantly different trajectory. In future work we want to show that this approach leads to robust robot action prototypes which work even in changing environments (i.e., step size).
\section{Discrete vs Continuous effect spaces}
\begin{figure} [h]
    \centering
    \begin{subfigure}[b]{.49\textwidth}
         \centering
         \includegraphics[width=\textwidth]{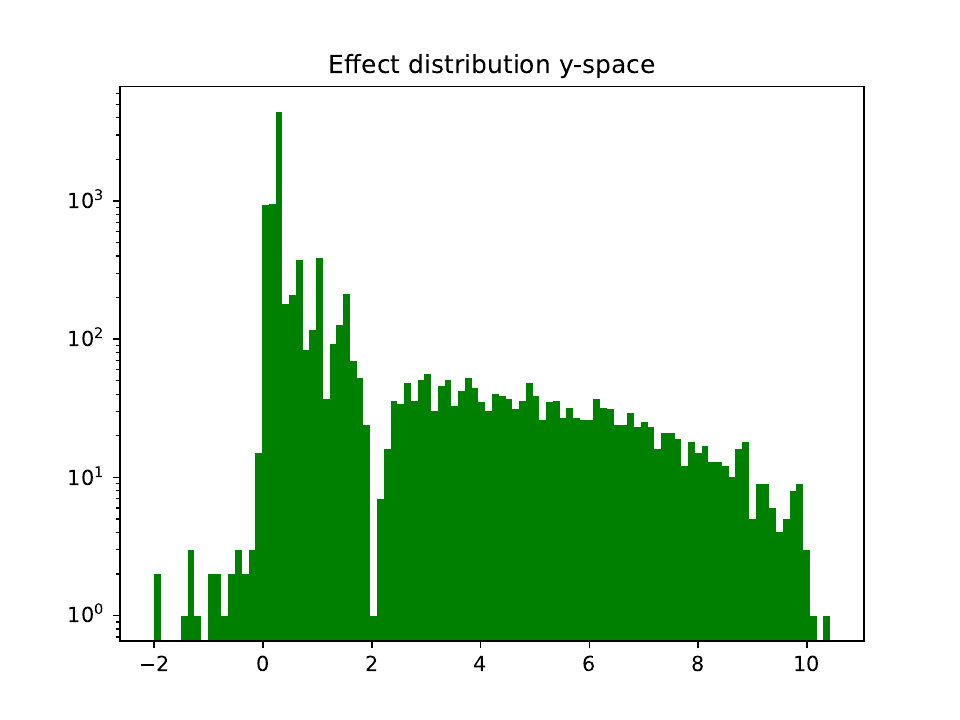}
         \caption{Continuous effect space distribution}
         \label{fig:continuous}
     \end{subfigure}
     \hfill
     \begin{subfigure}[b]{0.49\textwidth}
         \centering
         \includegraphics[width=\textwidth]{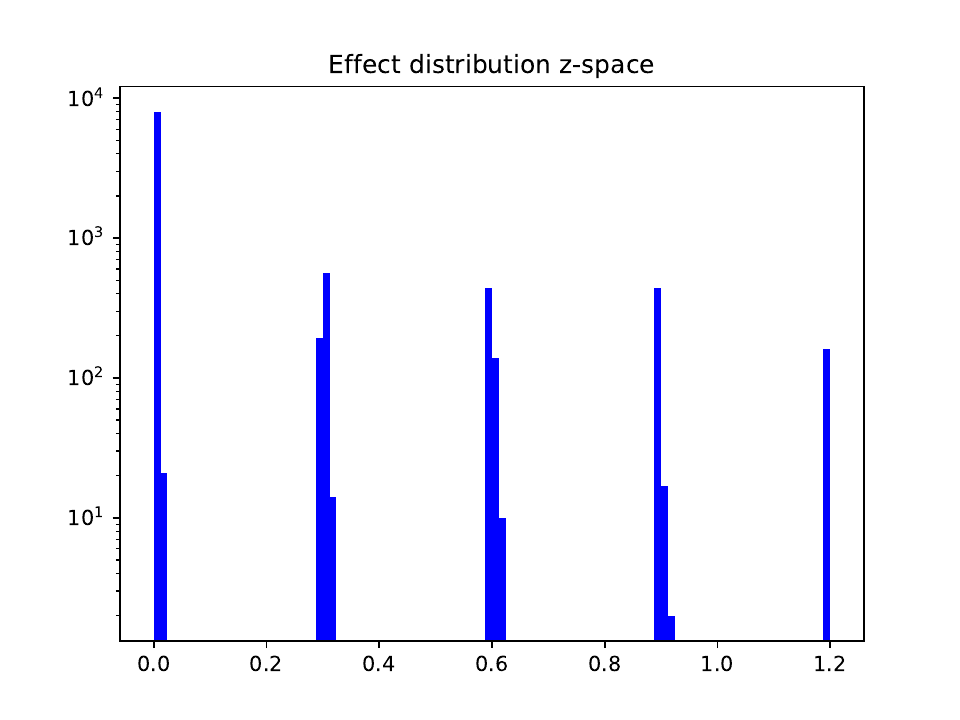}
         \caption{Discrete effect space distribution}
         \label{fig:discrete}
     \end{subfigure}
    \caption{Figure (a) shows the distribution of a continuous effect space (y-space of "Up the stairs" environment). Figure (b) displays the discrete effect space (z-space of "Up the stairs" environment). The different levels of height are clearly visible in the distribution (ground, first step, second step, etc.)}
    \label{fig:cont_discr_eff}
\end{figure}

This section provides a visual aid for understanding the difference between continuous and discrete effect spaces. Effects in the z direction are restricted by gravity. Each motion results in gravity pulling the robot back to a certain level, which manifests in clearly subdivided effect regions in effect space shown in figure \ref{fig:discrete}. Effects in the y direction are way less restricted, leading to continual distribution with no breaks as in figure \ref{fig:continuous}. While one on hand, the clustering classes in the discrete effect space are emergent, they are not apparent in the continuous case and, therefore, more susceptible to biases introduced by the clustering algorithm.
\section{Multi feature effect categories}
\begin{figure} [h]
    \centering
    \begin{subfigure}[b]{.49\textwidth}
         \centering
         \includegraphics[width=\textwidth]{img/effect_classes.pdf}
         \caption{y and z effect clustering}
         \label{fig:start}
     \end{subfigure}
     \hfill
     \begin{subfigure}[b]{0.49\textwidth}
         \centering
         \includegraphics[width=\textwidth]{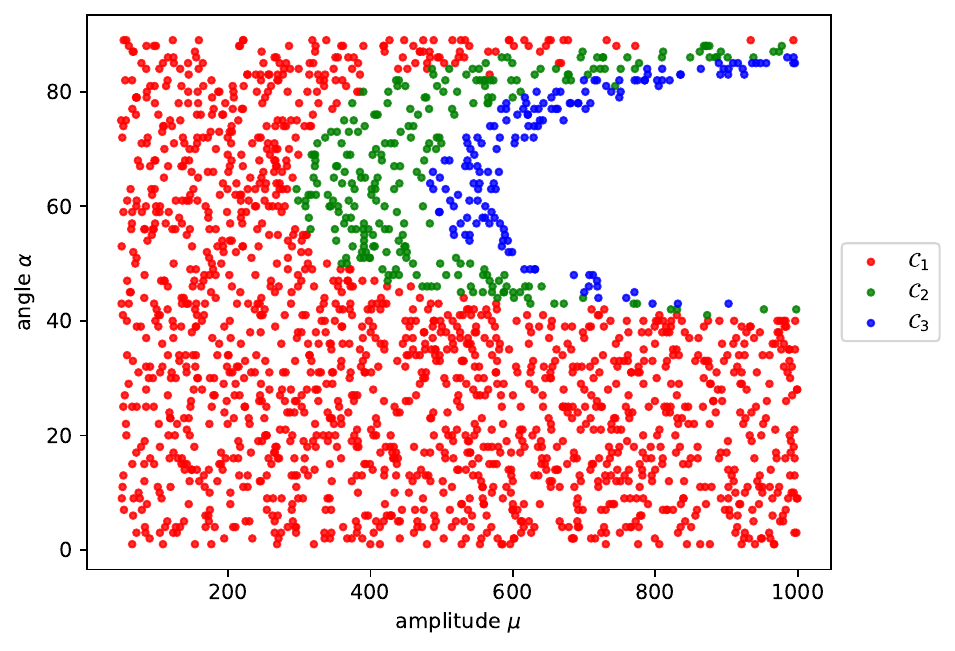}
         \caption{x, y, z effect clustering}
         \label{fig:multi-dim-eff}
     \end{subfigure}
    \caption{Emerging effect classes following section \ref{sec:methods} methodology for effect clustering. In both cases the clustering was done on the same effect sample collection. Figure (a) shows the effect categories when clustering on features y and z; Figure (b) shows the effect categories when clustering on features x, y and z.}
\end{figure}

This section illustrates the granularity-generalization trade-off when selecting the features to include in the effect categorization clustering (figure \ref{fig:multi-dim-eff}). In the "Up the stairs" environment the robot's motion capability is restricted to the y-z plane which makes any change in the x direction unintentional and therefore random. The addition of a random effect feature $x$ into the clustering process leads to the merging of the effect classes $\mathcal{C}_1$ with $\mathcal{C}_3$, $\mathcal{C}_2$ with $\mathcal{C}_5$ and $\mathcal{C}_4$ with $\mathcal{C}_6$.

\end{document}